\documentclass[letterpaper, 10 pt, conference]{ieeeconf}  % Comment this line out if you need a4paper

\IEEEoverridecommandlockouts                              % This command is only needed if 
\usepackage{graphicx}
\usepackage{subcaption}
\usepackage{booktabs}       % professional-quality tables
\usepackage{amsfonts, amssymb, amsmath}       % blackboard math symbols
\usepackage[capitalise]{cleveref}

\DeclareMathOperator*{\argmin}{arg\,min}

\newcommand{\myparagraph}[1]{\textbf{#1}~} %\noindent

\renewcommand{\eqref}[1]{(\ref{#1})}

\usepackage[backend=biber, maxbibnames=99, url=false, doi=false, isbn=false]{biblatex}
\usepackage{booktabs}

\def\method/{BiVO}  % this way we handle punctuation and get an error if / is forgotten.
\def\oracle/{NuScenes Oracle}
\def\dsonly/{DriverSensor}
\def\cvaeonly/{Trajectory CVAE}
\def\noreasoning/{Occlusion Agnostic}

\DeclareMathOperator*{\dkl}{D_{KL}}
\newcommand{\E}{\mathbb{E}}

\usepackage{xcolor}

\newcommand{\citet}[1]{\textcite{#1}}

\title{\LARGE \bf
Planning with Occluded Traffic Agents using \\ Bi-Level Variational Occlusion Models
}
\author{Filippos Christianos$^{1,3%,\text{\ddag}
}$, Peter Karkus$^{1}$, Boris Ivanovic$^{1}$, Stefano V. Albrecht$^{2,3}$ and Marco Pavone$^{1,4}$% "and' to add other authors <-this % stops a space
\thanks{$^{1}$NVIDIA Research, NVIDIA, %2788 San Tomas Expressway, 
Santa Clara, CA.
        {\tt\small \{pkarkus, bivanovic, mpavone\}@nvidia.com}}%
\thanks{$^{2}$Five AI / Bosch.
        {\tt\small stefano.albrecht@five.ai}}%
\thanks{$^{3}$ School of Informatics, University of Edinburgh.
        {\tt\small \{f.christianos, s.albrecht\}@ed.ac.uk}}%
\thanks{$^{4}$Department of Aeronautics and Astronautics, Stanford University.
        {\tt\small \{pavone@stanford.edu\}}}%
\thanks{%$^{\text{\ddag}}$ 
This work was done during an internship at NVIDIA. }%
}

\begin{document}

\maketitle
\thispagestyle{empty}
\pagestyle{empty}

%%%%%%%%%%%%%%%%%%%%%%%%%%%%%%%%%%%%%%%%%%%%%%%%%%%%%%%%%%%%%%%%%%%%%%%%%%%%%%%%
\begin{abstract}
Reasoning with occluded traffic agents is a significant open challenge for planning for autonomous vehicles. Recent deep learning models have shown impressive results for predicting occluded agents based on the behaviour of nearby visible agents; however, as we show in experiments, these models are difficult to integrate into downstream planning. 
To this end, we propose Bi-level Variational Occlusion Models (\method/), a two-step generative model that first predicts likely locations of occluded agents, and then generates likely trajectories for the occluded agents.  In contrast to existing methods, BiVO outputs a trajectory distribution which can then be sampled from and integrated into standard downstream planning. 
We evaluate the method in closed-loop replay simulation using the real-world nuScenes dataset. 
Our results suggest that \method/ can successfully learn to predict occluded agent trajectories, and these predictions lead to better subsequent motion plans in critical scenarios.
% We evaluate the use of these trajectories on a simple-sampling-based planning algorithm and present its performance on the real-world nuScenes dataset, including the hindsight cost. We also present qualitative examples of the trajectories generated by our method and how they affect the planner.

\end{abstract}

%%%%%%%%%%%%%%%%%%%%%%%%%%%%%%%%%%%%%%%%%%%%%%%%%%%%%%%%%%%%%%%%%%%%%%%%%%%%%%%%
\section{Introduction}

% Visibility during driving is far from perfect. Cars, trees, and buildings can obstruct the driver's view and only allow for limited observability of the state space. Humans often reason about hidden areas based on past experience and the behaviours of other drivers. 

\begin{figure}
    \centering
    \includegraphics{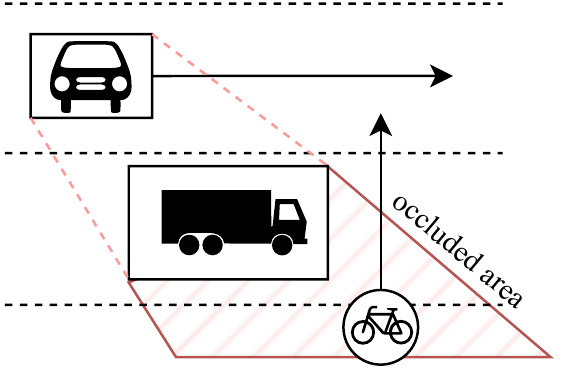}
    \caption{The ego vehicle is travelling with a constant speed and can observe a stopped truck on its right. The area behind the truck is occluded, and is hiding a bicycle that is attempting to cross the road. Both the ego vehicle and the bicycle are unaware of each other, leading to a dangerous situation.}
    \label{fig:intro_example}
\end{figure}

Reasoning with occluded traffic agents is an important open challenge for planning for autonomous vehicles. 
Planning under occlusions has an extensive literature in robotics; however, many prior works assume static occluded objects~\cite{callaghan2012gaussian,han2020planning}, or objects that are already detected and become occluded only temporarily~\cite{dequaire2018deep,sung2020hmpo}. 
Urban driving requires reasoning with the most challenging type of occlusions involving \emph{dynamic} and \emph{previously undetected} objects, because traffic agents, such as vehicles, cyclists, pedestrians, may emerge from occluded areas potentially with a high velocity. An example is shown in \cref{fig:intro_example}.

Classical planning approaches that reason with dynamic undetected traffic agents are often based on maintaining bounds on a worst case scenario~\cite{nager2019what,zhang2021safe}; however, the worst case scenario results in prohibitively conservative plans for driving in dense urban traffic. More recently, data-driven methods have been proposed that learn to predict likely occluded traffic agents from data.  In particular, \textcite{masha2022people} proposed a data-driven method that uses ``people as sensors'', that is, it trains a variational model to predict possible occluded areas given the past trajectory of visible traffic agents. While the method showed promising results on real-world data, it only predicts occluded space likely occupied by an agent, but not the agents' dynamic state or possible future trajectory. For this reason, the model cannot be easily integrated into downstream planning with dynamic agents, which we will further highlight in our experiments.

We introduce the Bi-level Variational Occlusion model (\method/), a data-driven occlusion prediction model that allows downstream planning with dynamic, previously undetected traffic agents. In its first step \method/ follows \textcite{masha2022people}: it predicts a probabilistic occupancy grid map (OGM)~\cite{elfes1989using} that captures possible occluded agents using a Conditional Variational Autoencoder (CVAE)~\cite{kihyuk2015learning}, that is conditioned on the past trajectory of a visible traffic agent. The model predicts an OGM for each visible agent and then fuses them into a single global OGM.  In the second, critical stage, \method/ predicts a distribution over future trajectories of possibly occluded agents using a second CVAE model, which is conditioned on the global OGM and other features of the environment.

We integrate \method/ into a sampling-based planning algorithm~\cite{karkus2022diffstack} for autonomous driving. The planner samples a set of dynamically feasible trajectories for the ego-vehicle, and selects the most promising trajectory given a hand-crafted cost function. \method/ predictions enter the planner through a collision avoidance term in the cost function. Specifically, we sample a large number of trajectories from \method/ along with their probabilities, and calculate the expected collision cost, treating the predicted occluded agent trajectories the same way as predictions for visible traffic agents. 

To validate \method/, we use real-world trajectory data from the nuScenes Prediction dataset~\cite{nuscenes}, both for direct prediction metrics, open-loop planning metrics, and in a closed-loop replay simulation. As one would expect, occluded objects rarely affect the desired trajectories of our planner, but when they do, reasoning about occlusions significantly improves the plan quality; and \method/ is significantly more effective than alternative learned models that were not designed for planning (see \cref{sec:baselines}). 

In summary, the contributions of this paper are as follows:
\begin{itemize}
    \item We introduce a generative model, \method/, based on variational autoencoders that is able to produce trajectories of occluded vehicles.
    \item We integrate \method/ into a fast sampling-based planning algorithm and % compare it with various variations that do not use our \method/ model. We 
    evaluate it in open and closed-loop replay simulation with the real-world nuScenes dataset.
    \item We demonstrate that \method/ predictions integrated into planning leads to better motion plans in critical scenarios.
    % \item We also integrate the SOTA occlusion prediction model of \citet{masha2022people} into our planner and show that without an effective way of reasoning with dynamics it is insufficient to plan high quality trajectories.
    \item To the best of our knowledge we are the first to integrate a learned occlusion model with a planning algorithm for autonomous driving.
\end{itemize}

\section{Related Work}

Detecting and reasoning with occluded objects in robotics has an extensive literature~\cite{chandel2015occlusion,gilroy2019overcoming}. 
In the context of autonomous vehicles, occlusions can be of critical importance. Indeed, prior work has proposed various methods that predict and/or plan with occluded traffic agents.

% \TODO{are we the first to do planning with a learned (or deep learned) occlusion prediction model? If so we should highlight this, also in the abstract+intro, if not, we should discuss it here.}

% Multiple methods exist to track objects behind occlusions, use computer vision to handle partially occluded objects, or reason on the behaviours of neighbouring agents. 

\myparagraph{Planning with occluded agents:}
Planning algorithms that reason with occluded agents typically rely on handcrafted occlusion models. For example, \textcite{franciszek2018tackling} propose an approach to predicting the presence of a vehicle coming out of an occluded region and ensures the existence of a fail-safe manoeuvre. \textcite{wang2021reasoning} extend this work by eliminating some of the occluded traffic by reasoning about the history of occlusions. \textcite{zhang2021safe} propose a method of navigating through traffic with occluded regions by making sure a potentially hidden pursuer should never intersect with the set of possible inevitable collision states. \textcite{hanna2021interpretable} use a model-driven approach that infers a joint distribution over the state of the occluded areas and the goals of other vehicles, using the observed trajectories of the vehicles.

% In contrast to these approaches, we use a data-driven approach to learn from the real-world nuScenes dataset how agents typically behave when occluded.
In contrast to these hand-crafted approaches, we propose a data-driven approach that learns a model of occluded agents from real-world data. 

\myparagraph{Data-driven occlusion models:}
Learning based models for occluded object prediction include~\textcite{schulter2018learning,purkait2019seeing,han2020planning}. However these models make assumptions about static objects or environments which are not pertinent in urban driving. Some learned models can handle dynamic traffic agents. Notably, \textcite{masha2022people} use an autoencoder architecture to infer the surroundings of visible objects and later reconstruct them into occupancy grid maps that encode the probability of occupied areas in 2D space. 
%However, it is not straightforward to integrate approaches that make occupancy predictions of areas with planners that typically reason about the future trajectories of agents. Solely reasoning about areas lacks the information on how agents might emerge out of occlusions and interfere with the ego vehicle.
However, it is not straightforward to integrate approaches that make occupancy predictions of areas with existing planners, since the predictions lack the information on how agents might emerge out of occlusions and interfere with the ego vehicle.

In our work, we predict dynamic agents together with their possible future trajectories instead of only occluded areas. Our model's predictions are key to integration with existing downstream planners that make use of probabilistic predictions of future trajectories.

\begin{figure*}[t]
    \centering
    \includegraphics{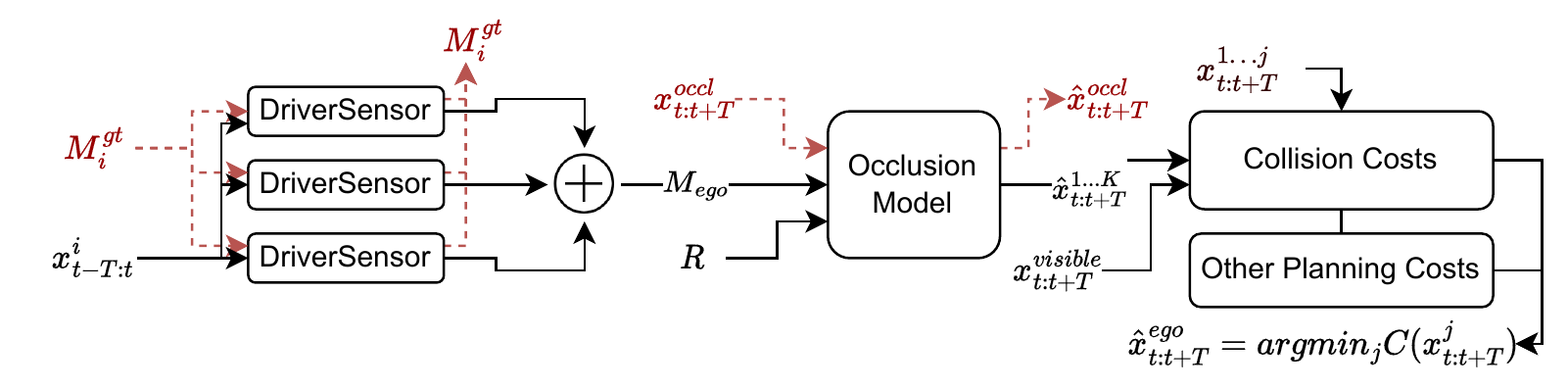}
    \caption{The information flow for \method/. The elements in red are only provided during training. During execution, the DriverSensor(s) reconstruct the surroundings of visible vehicles using only their observed trajectories. The OGMs are combined and given to a generative model which predicts plausible trajectories of occluded vehicles. The planner can then use that information to decide on a trajectory for the ego vehicle. }
    \label{fig:model}
\end{figure*}

\section{Technical Preliminaries}

\textbf{Variational Autoencoders:} Variational autoencoders~\cite{kingma_auto-encoding_2014} (VAEs) are generative models that aim to learn a density function over some unobserved latent variables $Z$ given a dataset input $x\in X$. Given an unknown true posterior $p(z|x)$, VAEs approximate it with a parametric distribution $q_\theta(z|x)$. The KL-divergence from the parametric distribution to the true posterior can be computed using:
\begin{multline*}
    \dkl(q_\theta(z|x) \Vert p(z|x)) =  \log p(x) \\
     - \E_{z\sim q_\theta(z|x)}[\log p_u (x|z)] + \dkl(q_\theta(z|x) \Vert q(z)),
\end{multline*}

where $\dkl$ is the KL divergence between two distributions, and the log-evidence term $\log p(x)$ is constant. The expectation and the KL-divergence (second line) are commonly called the negative evidence lower bound (ELBO). Minimising the ELBO is equivalent to minimising the KL-divergence between the parametric and the true posterior.

\textbf{States and trajectories:} A state $s^i_t$ for a vehicle $i$ is defined as the location, heading, velocity, and acceleration at the current timestep $t$. A trajectory $x^i_{t:t+T}$ is a sequence of states \(s^i_{t}, s^i_{t+1}, \dots, s^i_{t+T}\) that defines how an agent $i$ moved in time $T$. 

\textbf{Agents:} We will refer to the controlled vehicle as ``ego''. Other vehicles that are not controlled by the planner, pedestrians, or other road users will be referred to as ``agents''. Agents can be visible by the ego if they are in the line of sight, or occluded if there is an obstacle blocking their view (further details of this calculation is in \cref{sec:line-of-sight}).

\textbf{Occupancy grid maps:} OGMs encode the occupancy of an area. \(M_{i}^{\text{obs}} \in [0, 1]^{H, W}\) is the $H\times W$ area surrounding agent $i$ in a $1\times 1$ meter resolution, and each grid cell contains $1$ if it is occupied or $0$ if free. Locations that are not visible with a direct line of sight from the position of vehicle $i$ are marked as occluded with a value $0.5$. \(M_{i}^{\text{gt}}\) is the ground truth occupancy map of the same area.

\section{Bi-Level Variational Occlusion Models}\label{sec:methodology}
The objective of our occlusion model is to generate likely trajectories for agents emerging from occluded regions,  given a known map of occluded regions, the past and present state of visible agents, and a lane graph.

Our approach, \method/, is shown in \cref{fig:model}. We break down the problem into two subproblems and train separate CVAE models for each. Intuitively, the first step locates the subspace of occluded areas that have high potential of hidden objects; and the second step infers how these hidden object may emerge from the occluded space. 
%\method/ uses a hierarchical CVAE architecture, one CVAE model that locates occluded agents, and another CVAE that generates plausible trajectories conditioned on the previously predicted occluded agent locations.
Overall, \method/ parameterizes a distribution over trajectories that start from known occluded regions, and allows fast sampling from this distribution for subsequent planning.

\subsection{Reasoning about the Behaviour of Others}\label{sec:methodology:ogm}

\begin{figure}
    \centering
    \includegraphics{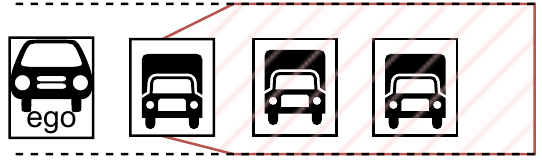}
    \caption{Illustrative example for using people as sensors. The ego vehicle can only observe the truck directly in front of it, which has stopped without any obvious reason. Even if the ego vehicle cannot directly observe the other vehicles (under the red striped area), it can assign a high probability mass on the possibility of an obstacle blocking the truck's way, just by reasoning about the stopped truck.}
    \label{fig:stopped_truck}
\end{figure}

The first component of our architecture, the DriverSensor, is based on the work of \textcite{masha2022people}. The DriverSensor model aims to produce OGMs with probabilities of obstacles existing in the respective grid cells (e.g. \cref{fig:stopped_truck}). The DriverSensor's CVAE is trained to reconstruct the OGMs of the immediate surroundings of visible cars by only observing their state (past locations, velocities, and accelerations). These reconstructions are later combined  into a single, larger OGM centered on the ego vehicle using the theory of belief functions (or Dempster-Shafer theory)~\cite{dempster1967upper}. The reconstruction \(M_{\text{ego}} \in [0,1]^{H, W}\) is an $H, W$ grid map that contains the probabilities of each cell being occupied. 

We use a similar architecture and loss function to \textcite{masha2022people} with a key modification as follows. The DriverSensor CVAE with discrete latent space consists of encoders \(p(z|x^i_{t-T:t})\) and \(q(z|x^i_{t-T:t},M_i^{\text{gt}}) \); and decoder \(p(M_i^{\text{gt}}|z)\). During training we minimise the loss $\mathcal{L}_{DS}$:

\begin{multline*}
    \mathcal{L}_{\text{DS}} = -\E_{z\sim q(z|x_{t-T:t})}\log p(M_i^{\text{gt}}|z) \\
     + \beta\dkl(q(z|x^i_{t-T:t},M_i^{\text{gt}}) \Vert p(z|x^i_{t-T:t})) \\
     - \mathcal{I}_{q}(M_i^{\text{gt}}; z) + (1-\beta)\mathcal{H}(q(\cdot|x^i_{t-T:t},M_i^{\text{gt}})),
\end{multline*}

where $\mathcal{I}$ is the mutual information and $\beta$ is a hyperparameter annealed from zero to one. We added the entropy term $\mathcal{H}$, calculated over the batch,  that maximises the number of active discrete latent classes in the beginning of training. We found this to be important to avoid the collapse of the discrete latent space.

%We also applied the Dempster-Shafer rule in the moving frame of reference of the ego vehicle, in contrast to the original work which uses the INTERACTION~\cite{interactiondataset} dataset that has a stationary frame of reference. 

In the rest of the document, we denote by $M_{\text{ego}}$ the most likely reconstruction of the fused DriverSensors. More details on the DriverSensor can be found in the original work \cite{masha2022people}.

%There are ways to infer the existence of hidden vehicles in occluded spaces\cite{}. For instance, observing the movement of vehicles and any changes to their velocities can inform of the possibility of occluded objects (e.g. \cref{fig:stopped_truck})

\subsection{Occluded Trajectory Generation}

An occupancy grid map, even the ground truth OGM, cannot be used to extrapolate the movement of hidden vehicles into the future. Our goal is to generate probable future trajectories of vehicles emerging from an occluded area. These trajectories can then be used as a component in a planning system. To model vehicles emerging from the occluded areas, we use the generative properties of a variational autoencoder. By doing so, we aim to learn from real-world data a distribution of vehicle trajectories that emerge from occluded areas.

The occlusion model consists of a CVAE that receives an occluded trajectory $x^{\text{occl}}_{t:T}$ as an input to the encoder. 
An ``occluded trajectory'' is defined as a trajectory that originates in an occluded area of the ego OGM (i.e. $M^{\text{obs}}_{\text{ego}} = 0.5$ at the location of the agent $i$ at time $t$).
The decoder is further conditioned on the latent variable $z$, a raster of the road layout $R$, and finally the most-likely reconstruction of the nearby occupancy grid map $M_{\text{ego}}$.
% At each timestep $t$, we denote an occluded trajectory $x^{\text{occl}}_{t:T}$ as one that originates in an occluded area of the ego OGM (i.e. $M^{\text{obs}}_{\text{ego}} = 0.5$ at the location of the agent $i$ at time $t$).

Formally, the occluded trajectories are projected into a latent space through the posteriors $q(z|x^{\text{occl}}_{t:T})$ and $p(z|x^{\text{occl}}_{t:T}, R, M_{\text{ego}})$. To learn the distribution, we optimise the objective $D_{KL}(q(z|x^{\text{occl}}_{t:T})||p(z|x^{\text{occl}}_{t:T}, R, M_{\text{ego}})$, where $D_{KL}$ is the KL divergence between two distributions. The evidence lower bound (ELBO) becomes:
\begin{multline}\label{eq:elbo}
    \mathcal{L}_{\text{Gen}} = -\E_{z\sim q(z|x^{\text{occl}}_{t:t+T})}\log p(x^{\text{occl}}_{t:t+T}|R, M_{\text{ego}}) \\ + D_{KL}(q(z|x^{\text{occl}}_{t:t+T}, R, M_{\text{ego}}) || p(z|R, M_{\text{ego}}))
\end{multline}
    
% Therefore, we assume that occluded areas in the field of vision of the ego vehicle hide a distribution from which we can sample trajectories. By using the VAE framework, we will learn parameters to approximate, and later sample, from this distribution. To improve our approximation of the underlying distribution, we also inform the decoder of additional information. Our conditional VAE uses a decoder that is conditioned on the latent variable $z$, a raster of the road layout $R$, and finally the most-likely reconstruction of the nearby occupancy grid map $M_{\text{ego}}$.

Therefore, the occluded areas in the field of vision of the ego vehicle hide a distribution from which we can sample trajectories. After training the generative model on real-world data, we sample $z\sim N(0,1)$ which can then be decoded as trajectories using the VAE's decoder.

\section{Planning with BiVO}

The generated trajectories can be directly integrated into a planning component. Our planner follows a common approach for AV planning, where we first generate a set of candidate trajectories, and then choose the most promising trajectory based on a hand-crafted cost function. 

More specifically, we use the planning algorithm of \citet{karkus2022diffstack}. The algorithm generates $J$ \emph{dynamically feasible} candidate trajectories \(x^{1 \dots j}_{t:T}\), by first sampling terminal points based on the lane map, connecting the current state to the terminal point through spline interpolation, and filtering the trajectories that violate control limits. We then evaluate each of these trajectories using handcrafted cost components, specifically

\begin{equation}\label{eq:planning}
    C = C_{\text{hd}} + C_{\text{vd}} + C_{\text{ef}} + C_{\text{col}} + C_{\text{goal}},
\end{equation}

where $C_{\text{vd}}$ and $C_{\text{hd}}$ are the lane and heading deviation respectively, calculated from the nearest lane at each spline point. $C_{\text{ef}}$ is the effort cost, calculated as the acceleration squared and $C_{\text{goal}}$ is the distance from the goal. In experiments we simply define the goal as the ground truth future state at time $t+T$. 

Finally, $C_{\text{col}}$ is the collision cost, calculated using the trajectories with the visible but also with the predicted vehicles. We define this collision cost using:

\begin{equation*}
    C_{\text{col}} = \sum_{i=1}^{N}{\sum_{\tau=t}^{T}{\phi(s^i_\tau - s^{\text{ego}}_\tau)}} + \sum_{k=1}^{K}{\pi_k\sum_{\tau=t}^{T}{\phi(s^k_\tau - s^{\text{ego}}_\tau)}} ,
\end{equation*}

where $i$ iterates over the $N$ visible agents (we use ground truth future trajectories for simplicity), $k$ iterates over the $K$ trajectories sampled by our method, and $\phi$ is a radial basis function. \(\pi_k\) represents the probability of trajectory $k$ and is defined as $\frac{1}{K} \pi_e$ with $\pi_e$ being the prior for the existence of any occluded agent.

Eventually, to find the desired ego trajectory, the planner calculates:
\begin{equation}\label{eq:planning_min}
     \hat x^{\text{ego}}_{t:t+T} = \argmin_j C( x^{j}_{t:t+T})
\end{equation}

\section{Experiments}

We evaluate \method/ using the nuScenes dataset~\cite{nuscenes}. We aim to evaluate i) the quality of generated trajectories, and ii) how planning is affected when using these trajectories. 

% The quality of the trajectories could be evaluated using the marginal likelihood of the data under the model, which however is intractable. Instead, in \cref{tab:loss}, we report the ELBO loss, i.e. a lower bound of the marginal likelihood. In \cref{sec:exp:quality} we also provide some qualitative examples. 
%We also report the open-loop and closed-loop hindsight costs of selected trajectories for various planner setups, including \method/.

\subsection{Data Processing \& Experimental Details}\label{sec:line-of-sight}
\begin{figure}
    \includegraphics[trim={2.5cm 5.5cm 2.5cm 0cm},clip,width=\columnwidth]{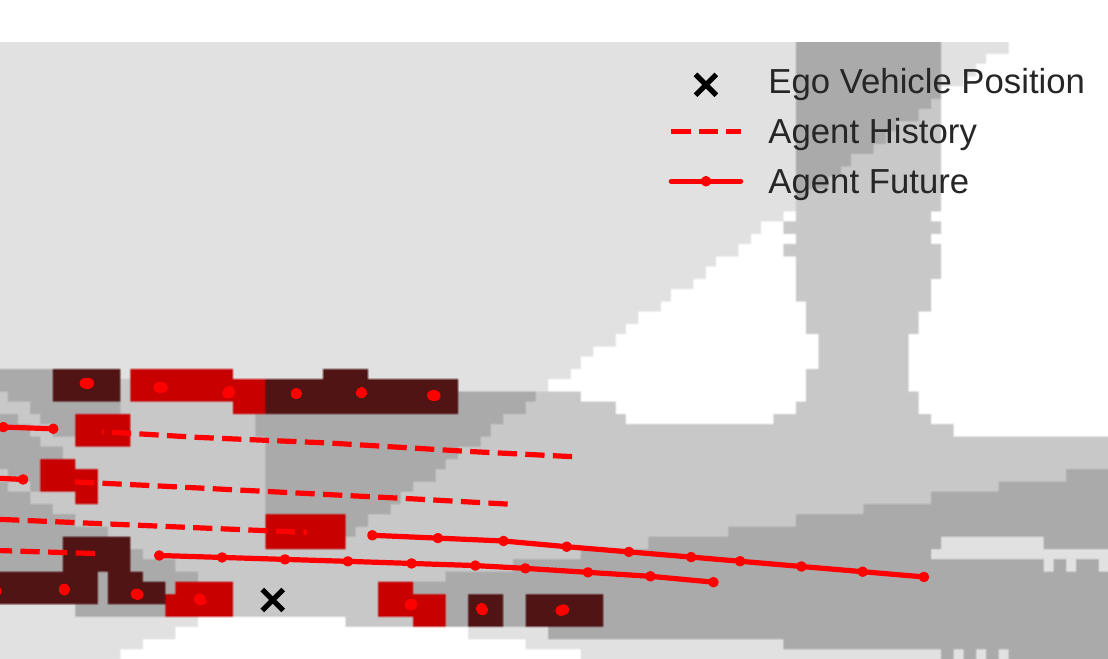}
    \caption{Line of sight calculation starting from the ego vehicle position. Agents that are in a direct line of sight (calculated from the X location with a 1x1m resolution) can be observed by the ego agent (bright red). Areas that are not in a line of sight are denoted by a grey shadow, including the dark red agents.}\label{fig:lineofsight}
\end{figure}

The original nuScenes dataset \cite{nuscenes} does not contain occlusion information, therefore we calculate at every timestep the occlusion using a direct line-of-sight method. Specifically, we first create a raster of the surrounding scene that includes the neighbour agents in a $1\times1$ meter grid. We draw lines starting at the centre of the ego vehicle towards the edges of the map, which however stop when an agent is encountered. Any area covered by those lines is considered visible, otherwise it is considered occluded (see \cref{fig:lineofsight}). During planning, we only use the trajectories of visible vehicles (except in the \oracle/ baseline). 

\begin{table}[t]
\centering
\begin{tabular}{@{}lll@{}}
\toprule
           & Training $\downarrow$                     & Test $\downarrow$                     \\ \midrule
\method/       & $38.82 \pm 0.42$ & $48.68 \pm 0.61$ \\
\cvaeonly/       & $45.98 \pm 0.13$ & $52.71 \pm 0.10$ \\
Simple CVAE & $68.32 \pm 0.06$ & $73.52 \pm 0.06$ \\ \bottomrule
\end{tabular}\caption{ELBO at the end of training for both training and test sets. Mean and standard error over three training seeds.}\label{tab:loss}
\end{table}

To train our models, we first independently train the DriverSensor model on the nuScenes data to produce the \(M_i\) OGMs which are then merged into \(M_{\text{ego}}\). After the DriverSensor model is trained, its output is used to train our VAE model by minimising the ELBO loss (\cref{eq:elbo}). 

%\Cref{eq:planning} is used to select trajectories in an open-loop and closed-loop evaluation. We report the hindsight cost using the ground truth occluded vehicles, i.e. the trajectory that minimises \cref{eq:planning_min} is first calculated, then the same cost function is reevaluated but now uses all the ground truth trajectories of surrounding vehicles. 

For DriverSensor, we used the hyperparameters specified in \textcite{masha2022people}. For our occlusion model, we set the learning rate to $3e-4$, and we used four linear layers for encoding and decoding the trajectories. We used the VQ-VAE~\cite{oord2017vqvae} for our autoencoder architecture. We trained for a total of five epochs on the nuScenes dataset, on timesteps that have at least $T$ seconds of history and future. Finally we used $T=5s$ as a time horizon for both planning and trajectory reconstruction. Any generated trajectories that are not feasible or begin on a non-occluded area are filtered out. 
% For \cref{tab:loss} we used three seeds.

\begin{table*}[t]
\centering
\begin{tabular}{@{}lccc@{}}
\toprule
                                            & Open loop $\downarrow$ (all scenes) & Open loop $\downarrow$ (critical scenes) & Closed loop $\downarrow$ (all scenes)\\ 
\midrule
NuScenes Oracle           &   4.4922                 &   5.7410                   &   0.3502            \\ 
\midrule
\method/ (ours)                              &    \textbf{4.8932} (+8.92\%)   &   \textbf{6.0843} (+5.97\%)    &   \textbf{0.3472} (-0.85\%)  \\

Trajectory CVAE                 &    5.2639 (+17.17\%)     &   6.2254 (+8.43\%)         &   0.3479 (-0.65\%)      \\
DriverSensor \cite{masha2022people}                   &    5.1949 (+15.64\%)     &   6.3268 (+10.20\%)        &   0.3532 (+0.85\%)      \\
\midrule
Occlusion agnostic baseline             &    4.5062 (+0.31\%)      &   6.2021  (+8.03\%)        &   0.3503 (+0.02\%)     \\  
\bottomrule
%Ground Truth Trajectories                   &   7.850194                 &   11.495206                  &   --           \\ 
\end{tabular}
\caption{Average hindsight cost for our method and various baselines in the nuScenes dataset (lower is better). In parenthesis we denote the percentage increase of cost over the \oracle/.}\label{tab:openloop}
\end{table*}

% \begin{table*}[t]
% \centering
% \begin{tabular}{@{}llll@{}}
% \toprule
%                                             & Open Loop Hindsight Cost (Full) & Open Loop Hindsight Cost (Critical Scenes) & Closed Loop Hindsight Cost\\ \midrule
% \method/ (ours)                              &    4.8932                &   6.0843                   &   0.3472   \\
% \noreasoning/              &    4.5062                &   6.2021                   &   0.3503       \\  
% \cvaeonly/                &    5.2639                &   6.2254                   &   0.3479       \\
% \dsonly/                   &    5.1949                &   6.3268                   &   0.3532      \\
% \midrule
% \oracle/           &   4.4922                 &   5.7410                   &   0.3502            \\ \bottomrule
% %Ground Truth Trajectories                   &   7.850194                 &   11.495206                  &   --           \\ 
% \end{tabular}
% \caption{Average hindsight cost for our method and various baselines. Contains all scenes starting at $T$ seconds of the nuScenes validation set. Closed loop scenes were concatenated to 15 seconds. Lower is better.}\label{tab:openloop}
% \end{table*}

\subsection{Baselines}\label{sec:baselines}

We compare \method/ with the baselines described below.

\textbf{\noreasoning/:} This baseline assumes that occluded spaces do not contain any occluded agents and should not affect the trajectory of the ego vehicle. This is equivalent to setting assigning a zero probability to any generated trajectories ($\pi_k = 0$).

\textbf{\cvaeonly/:} Uses a CVAE in a similar configuration to our method. However, it does not make use of a reconstructed OGM $M_{\text{ego}}$ which acts as a prior to the trajectory generation process. To keep the number of parameters and architecture configuration fair, we instead use the an OGM that notates the locations that are occluded (i.e. $M^{\text{obs}}_{\text{ego}}$, see \cref{sec:methodology:ogm}).

\textbf{\dsonly/:} The SOTA occlusion model of \textcite{masha2022people} that outputs an OGM instead of trajectories. To integrate the model with planning, we generate agents in a heuristic manner, selecting the highest-valued location(s) in $M_{\text{ego}}$ and generating occluded vehicles there. While this method could potentially be better at locating the initial location of occluded agents, it does not learn how these agents behave or affect the ego vehicle.

\textbf{\oracle/:} The trajectories of occluded vehicles are fully observed and used by the calculations of this baseline. The collision cost of \cref{eq:planning} uses all nearby agents for evaluating \cref{eq:planning_min}. Notably, this baseline is prone to errors in the dataset, and while we do not explicitly remove vehicles not in the line of sight, the dataset itself could contain such occlusions (from trucks, walls, objects, or even sensor failures). However, in the open-loop evaluation this baseline acts as a lower bound to the minimised cost since it uses the same data for calculation of the hindsight cost.

%\textbf{Ground Truth:} Ground truth trajectory as selected by the human driver. The human driver does not directly optimise \cref{eq:planning_min} and therefore has a higher hindsight cost. This does not mean that the trajectory of the driver is inferior to our planner, just that there might be more than one way to evaluate candidate trajectories.

\subsection{Experimental Results}

\begin{figure*}[t]
\centering
    \begin{subfigure}{0.249\textwidth}
      \centering
      \includegraphics[trim={2.5cm 5.5cm 2.5cm 0cm},clip,width=\linewidth]{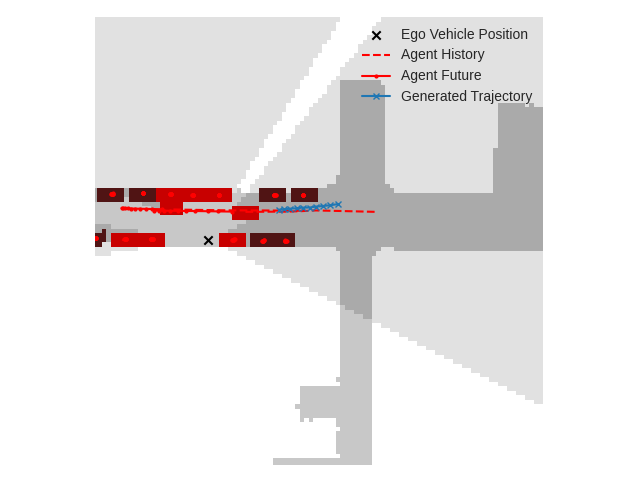}
      \caption{An agent driving straight\\ in an occluded space.}
      \label{fig:traj:1}
    \end{subfigure}%
    \begin{subfigure}{0.249\textwidth}
      \centering
      \includegraphics[trim={2.5cm 5.5cm 2.5cm 0cm},clip,width=\linewidth]{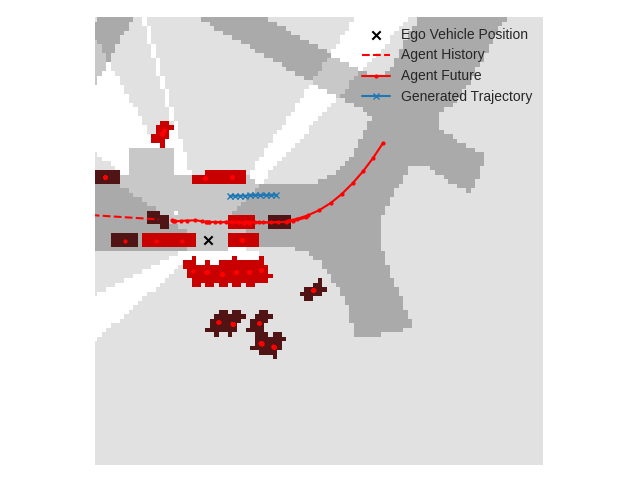}
      \caption{An oncoming traffic agent\\in the opposing side of the road.}
      \label{fig:traj:2}
    \end{subfigure}%
    \begin{subfigure}{.249\textwidth}
      \centering
      \includegraphics[trim={2.5cm 5.5cm 2.5cm 0cm},clip,width=\linewidth]{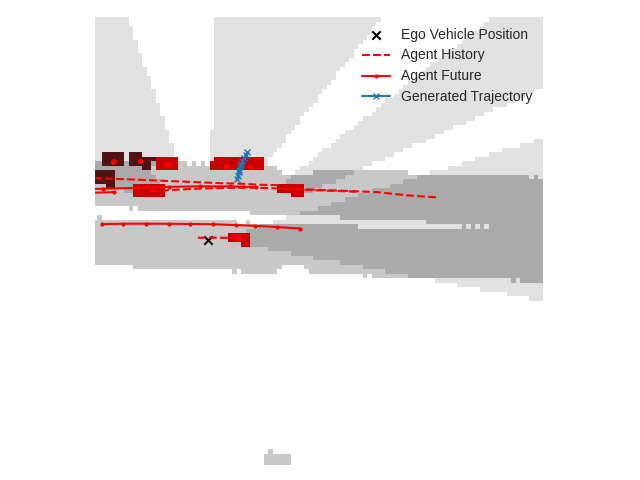}
      \caption{A pedestrian appearing\\behind parked vehicles.}
      \label{fig:traj:3}
    \end{subfigure}%
    \begin{subfigure}{.249\textwidth}
      \centering
      \includegraphics[trim={2.5cm 5.5cm 2.5cm 0cm},clip,width=\linewidth]{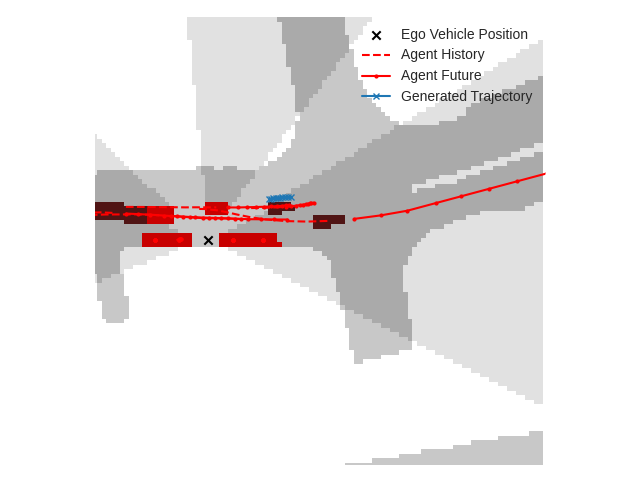}
      \caption{A generated trajectory next to an actual occluded agent.}
      \label{fig:traj:4}
    \end{subfigure}%
    \caption{Examples of trajectories (in blue) generated by \method/. Darker areas are occluded, bright red cars are visible.}\label{fig:trajs}
\end{figure*}

Our final qualitative results are reported in \cref{tab:loss,tab:openloop} and our qualitative results are in \cref{fig:manual-example,fig:trajs}. In this section, we provide a further analysis of these results.

\Cref{tab:loss} shows the ELBO loss as a proxy to the marginal likelihood of \(x^{\text{occl}}_{t:t+T}\) under the model. \method/ achieves a lower ELBO for both the training and the test set. We also compared with a simple CVAE model that does not use either \(M_{\text{ego}}\) or \(M^{\text{obs}}_{\text{ego}}\). These results show that information about the occluded areas is important, and further that our bi-level approach can make use of the information inferred by the DriverSensors.

Our planning experiments show that occluded agents do not often interfere with the ego trajectory. Specifically, in $98\%$ of the scenes in the nuScenes test set, the \oracle/ and the \noreasoning/ baselines select identical trajectories. This is not surprising, as many of the scenes contain relatively simple and safe scenarios, such as driving on a straight path. Indeed, much of the driving that happens in a normal situation does not deal with stray agents as motivated in \cref{fig:intro_example}, but instead any agents come in the field of view of the ego vehicle long before any preventive action needs to be taken. That said, our algorithms must consider such situations even if they are rare in nature. 

\Cref{tab:openloop} shows the mean \emph{hindsight cost} of the selected trajectories when planning with different occlusion models. We report the hindsight cost using the ground truth occluded vehicles, i.e. \(\hat x^{\text{ego}}_{t:t+T}\) using \cref{eq:planning_min} is first calculated, then the same cost function is reevaluated but now using the ground truth trajectories. So, the hindsight cost is the cost of the selected trajectory evaluated under the fully observable current and future states (i.e. the cost function of \oracle/). The open-loop results are split into two sets: the first set contains the full nuScenes evaluation dataset; the second set contains the ``Critical Scenes'' defined as any scenes where there is at least one timestep where the cost of the best trajectory selected by the \noreasoning/ and \oracle/ baselines are different.

\textbf{Open loop (all scenes):} In the full validation dataset, the \noreasoning/ baseline performs very closely to the fully observable (\oracle/) lower bound. This is largely a result of the simpler scenes that do not contain any important occluded agents. \method/ selects trajectories with lower hindsight cost ($+8.92\%$ increase over the \oracle/) than both \cvaeonly/ ($+17.17\%$) and the \dsonly/ ($+15.64\%$).

\textbf{Open Loop (critical scenes):} In the critical scenes, where occlusions do make a difference in the decision making process, our method has lower hindsight cost than all baselines, including \noreasoning/. Specifically, \method/ only has a $+5.97\%$ increase over the \oracle/ with \cvaeonly/ and \dsonly/ having $+8.43\%$ and $+10.20\%$ respectively. The increased hindsight cost of \noreasoning/  ($+8.03\%$) indicates an increased collision cost over \method/.
Therefore, in the critical scenes, our experiments show that our method selects trajectories that consider the existence of occluded agents that might interfere with the ego vehicle and makes better trajectory decisions.

\textbf{Closed loop (all scenes):} In the closed-loop evaluation (right-most column in \Cref{tab:openloop}) we present the hindsight cost of the actual trajectory. \method/ selects trajectories with the lowest hindsight cost ($-0.85\%$). This suggests that \method/ trajectories are even better than the \oracle/, which is also the case for \cvaeonly/ ($-0.65\%$). One possible reason is that scenes in the nuScenes dataset may contain occlusions themselves. Occluded agents may be missing from the data, and in some situations such agents may appear suddenly interfering with the ego vehicle. Further, in our experiments non-ego agents follow their recorder ``ground truth'' trajectory, therefore they do not react to the ego's trajectory which may cause unreasonable collisions. To mitigate this issue we reduced the maximum time per scene to 15 seconds. \dsonly/ ($+0.85\%$) and \noreasoning/ ($+0.02\%$) still select trajectories with higher hindsight cost.

In conclusion, our experiments show that \method/ comes with the trade-off of slightly costlier trajectories ($+8.58\%$ over being agnostic reasoning about occlusions) in regular driving scenarios but offers an improvement ($-1.93\%$) in critical situations. As we show in \cref{sec:exp:quality} and \cref{fig:manual-example}, the selected trajectories are reasonable and lead to preferred states when occluded agents ultimately become visible.

%In conclusion, our experiments show that \method/ comes with the trade-off of more expensive in terms of hindsight costs trajectories when the occluded space does not hide any agents that would interfere with the ego's trajectories. However, this should not always be expected to be the case, and when such agents exist, \method/ significantly improves the position of the ego agent. 

In addition, \method/ can be used in real-time. In our experiments, inference only requires a few milliseconds, significantly less that the planner's frequency, depending on the quantity of visible neighbouring agents and the number of samples drawn from the generative model (we used 1000).

\subsection{Qualitative Results}\label{sec:exp:quality}

\begin{figure}
    \includegraphics[width=\columnwidth]{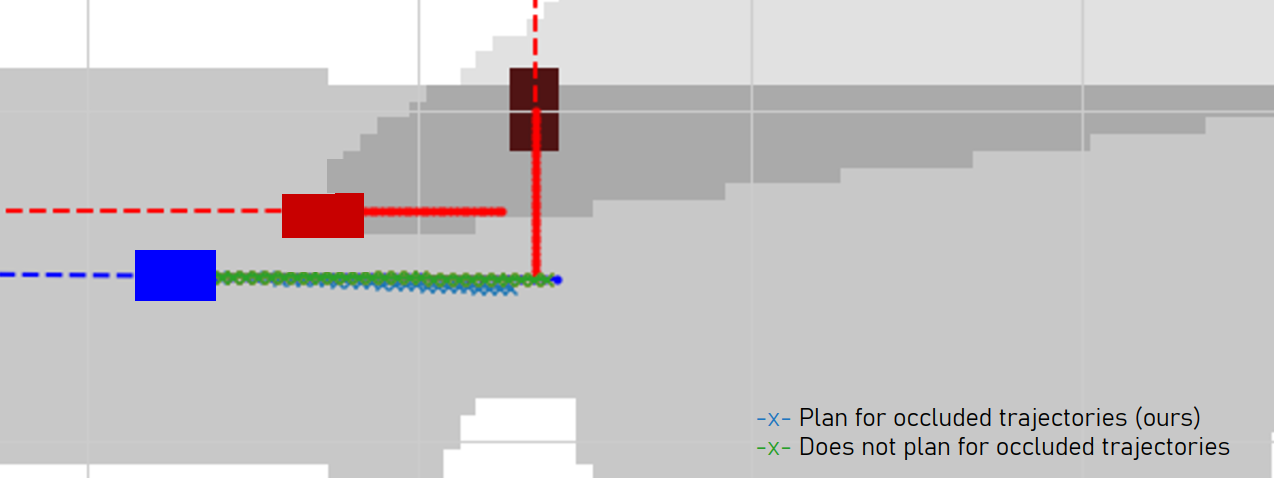}
    \caption{Reconstruction of example in \cref{fig:intro_example}. The ego vehicle (blue) cannot observe the occluded agent crossing the road. The (bright red) agent that is visible, has noticed the occluded agent and is breaking. We plot the trajectory selected by \noreasoning/ and our method. Notably, even if our method cannot observe the occluded agent, it chooses a trajectory slightly to the right and with lower speed.} \label{fig:manual-example}
\end{figure}

\textbf{Trajectories sampled from \method/:} In \cref{fig:trajs}, we present examples of trajectories drawn from \method/. The trajectories were generated by sampling $z\sim N(0, 1)$ once and combining them with the raster $R$ and the reconstruction $M^{\text{ego}}$. While in those figures we draw a single sample for clarity, during the planning experiments we sample up to 1000 trajectories per scene. 

Many of the trajectories drawn from \method/ are plausible for the respective occluded areas and road layouts. For instance, both \cref{fig:traj:2,fig:traj:3} show agents driving normally in their assigned lanes. This shows a benefit of our approach to learn from data: trajectories are often close to what would be encountered in the real world. \Cref{fig:traj:4} even generates a trajectory very close to a vehicle that indeed exists, but is hidden from the model and the planner. Finally, \cref{fig:traj:3} shows a more aggressive trajectory, a pedestrian who is emerging from two parked vehicles, a very dangerous situation in the real world, and one that must be accounted for (also see \cref{fig:manual-example}). 
It has to be noted that individual generated trajectories do not have a significant weight in the planner's calculation. Instead, the planner sums over hundreds of such trajectories to grasp the underlying distribution of plausible occluded agents. %\comment{cut for space}

\textbf{Planning in a dangerous situation:} To build an intuition of the selected trajectories of \method/, we have reconstructed the motivational example of \cref{fig:intro_example} in \cref{fig:manual-example}. The ego vehicle is depicted in blue, and is moving without being able to observe the occluded agent (dark red). The only information of the existence of any obstacle in the ego's path, would be the visible agent slowing down. However, even that might not be enough to provide evidence of any obstacle. Our method still decides on a trajectory that is slightly slower, but also slightly away from the occluded area.  While this is a worst-case scenario, a vehicle that uses \method/ and considers the learned distribution of occluded vehicles will i) have slower speed ($3.65 m/s$ vs $4m/s$) , ii) be further away (by $1.6m$), and iii) observe the occluded agent sooner than not reasoning about occlusions in its planning.

\section{Conclusion \& Future Work}

We introduced \method/, a bi-level variational autoencoder model to generate plausible agent trajectories in occluded areas. We trained this model in the real-world nuScenes dataset and integrated it with a simple sampling-based planner. The planner produces trajectories that take into consideration occlusions and while they have a possibly higher hindsight cost in simple driving scenarios, they do offer positional advantages in critical scenarios where occluded agents are indeed important to driving. In the future, we aim to improve our model by including realistic vehicle dynamics, and integrating previously observed information on occluded objects.

% \addtolength{\textheight}{-12cm} % This command serves to balance the column lengths
   % This command serves to balance the column lengths
                                  % on the last page of the document manually. It shortens
                                  % the textheight of the last page by a suitable amount.
                                  % This command does not take effect until the next page
                                  % so it should come on the page before the last. Make
                                  % sure that you do not shorten the textheight too much.

%%%%%%%%%%%%%%%%%%%%%%%%%%%%%%%%%%%%%%%%%%%%%%%%%%%%%%%%%%%%%%%%%%%%%%%%%%%%%%%%

%%%%%%%%%%%%%%%%%%%%%%%%%%%%%%%%%%%%%%%%%%%%%%%%%%%%%%%%%%%%%%%%%%%%%%%%%%%%%%%%

%%%%%%%%%%%%%%%%%%%%%%%%%%%%%%%%%%%%%%%%%%%%%%%%%%%%%%%%%%%%%%%%%%%%%%%%%%%%%%%%
% \section*{APPENDIX}

% Appendixes should appear before the acknowledgment.

% \section*{ACKNOWLEDGMENT}

% The preferred spelling of the word ÒacknowledgmentÓ in America is without an ÒeÓ after the ÒgÓ. Avoid the stilted expression, ÒOne of us (R. B. G.) thanks . . .Ó  Instead, try ÒR. B. G. thanksÓ. Put sponsor acknowledgments in the unnumbered footnote on the first page.

%%%%%%%%%%%%%%%%%%%%%%%%%%%%%%%%%%%%%%%%%%%%%%%%%%%%%%%%%%%%%%%%%%%%%%%%%%%%%%%%

% \begin{thebibliography}{99}
% \end{thebibliography}
\printbibliography
\end{document}